\documentclass{article}
\usepackage{ijcai24}

\usepackage{times}
\usepackage{soul}
\usepackage{url}
\usepackage[hidelinks]{hyperref}
\usepackage[utf8]{inputenc}
\usepackage[small]{caption}
\usepackage{graphicx}
\usepackage[inkscapelatex=false]{svg}
\usepackage{amsmath}
\usepackage{amsthm}
\usepackage{booktabs}
\usepackage{algorithm}
\usepackage{algorithmic}
\usepackage[switch]{lineno}

\usepackage{textcomp}
\usepackage{adjustbox}
\usepackage{booktabs} 
\usepackage{multirow} 
\usepackage{tabularx}
\usepackage{makecell}
\usepackage{array}

\title{\includegraphics[width=0.8cm]{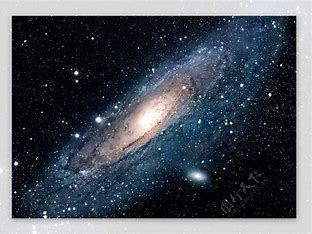} EmoVerse: Exploring Multimodal Large Language Models \\ for Sentiment and Emotion Understanding}

\author{
Ao Li$^{1,\dagger}$
\and
Longwei Xu$^{1,\dagger}$\and
Chen Ling$^2$\and
Jinghui Zhang$^1$\and
Pengwei Wang$^{1,*}$
\\
\affiliations
$^1$Shandong University
$^2$Xi'an Jiaotong University\\
}

\begin{document}
\maketitle
\begin{abstract}
Sentiment and emotion understanding are essential to applications such as human-computer interaction and depression detection. While Multimodal Large Language Models (MLLMs) demonstrate robust general capabilities, they face considerable challenges in the field of affective computing, particularly in detecting subtle facial expressions and handling complex emotion-related tasks, such as emotion reason inference and understanding emotions in long-context scenarios. Furthermore, there is a lack of a unified MLLM that can effectively handle both sentiment and emotion-related tasks. To address these challenges, we explore multi-task training strategies for MLLMs in affective computing and introduce \textbf{Emotion Universe (EmoVerse)}, an MLLM designed to handle a broad spectrum of sentiment and emotion-related tasks. In addition, EmoVerse is capable of deeply analyzing the underlying causes of emotional states. We also introduce the \textbf{Affective Multitask (AMT) Dataset}, which supports multimodal sentiment analysis, multimodal emotion recognition, facial expression recognition, emotion reason inference, and emotion cause-pair extraction tasks. Extensive experiments demonstrate that EmoVerse outperforms existing methods, achieving state-of-the-art results in sentiment and emotion-related tasks. The code is available at \url{https://github.com/liaolea/EmoVerse}.
\end{abstract}

{
\renewcommand{\thefootnote}%
{\fnsymbol{footnote}}
\footnotetext[0]{$\dagger$ Equal contribution. * Corresponding authors.} 
}

\section{Introduction}




Sentiment and emotion understanding are crucial for applications such as human-computer interaction and mental health monitoring. 
Traditional unimodal approaches, which rely solely on facial expressions~\cite{jiang2020dfew}, text~\cite{lei2023instructerc}, or audio~\cite{hsu2021hubert}, have made progress but remain inherently limited, as each modality captures only a partial view of human emotions, making it challenging to achieve a comprehensive understanding. 
To address these limitations, researchers~\cite{yang2023confede,li2023decoupled,lv2021progressive,ghosal2019dialoguegcn,hu2021dialoguecrn,hu2022mm,sun2021discourse,hu2021mmgcn,zhang2024reconstructing,cheng2023semi,wang2024incomplete} have increasingly turned to multimodal methods that integrate multiple modalities for sentiment analysis or emotion recognition. 
While these approaches have enhanced classification accuracy, they predominantly focus on identifying sentiment or emotions, rather than delving into their underlying causes or reasoning about emotional contexts. 
Recent multimodal large language models (MLLMs) have made significant strides in general visual-language understanding~\cite{liu2024visual}. 
However, in the field of affective computing, their zero-shot performance remains limited, particularly in tasks such as facial expression recognition, emotion reason inference, and contextual emotion understanding. 
Moreover, there is currently a lack of an MLLM capable of performing multiple emotion-related tasks. 

\begin{figure*}[h]
  \centering
  \includegraphics[width=1\linewidth]{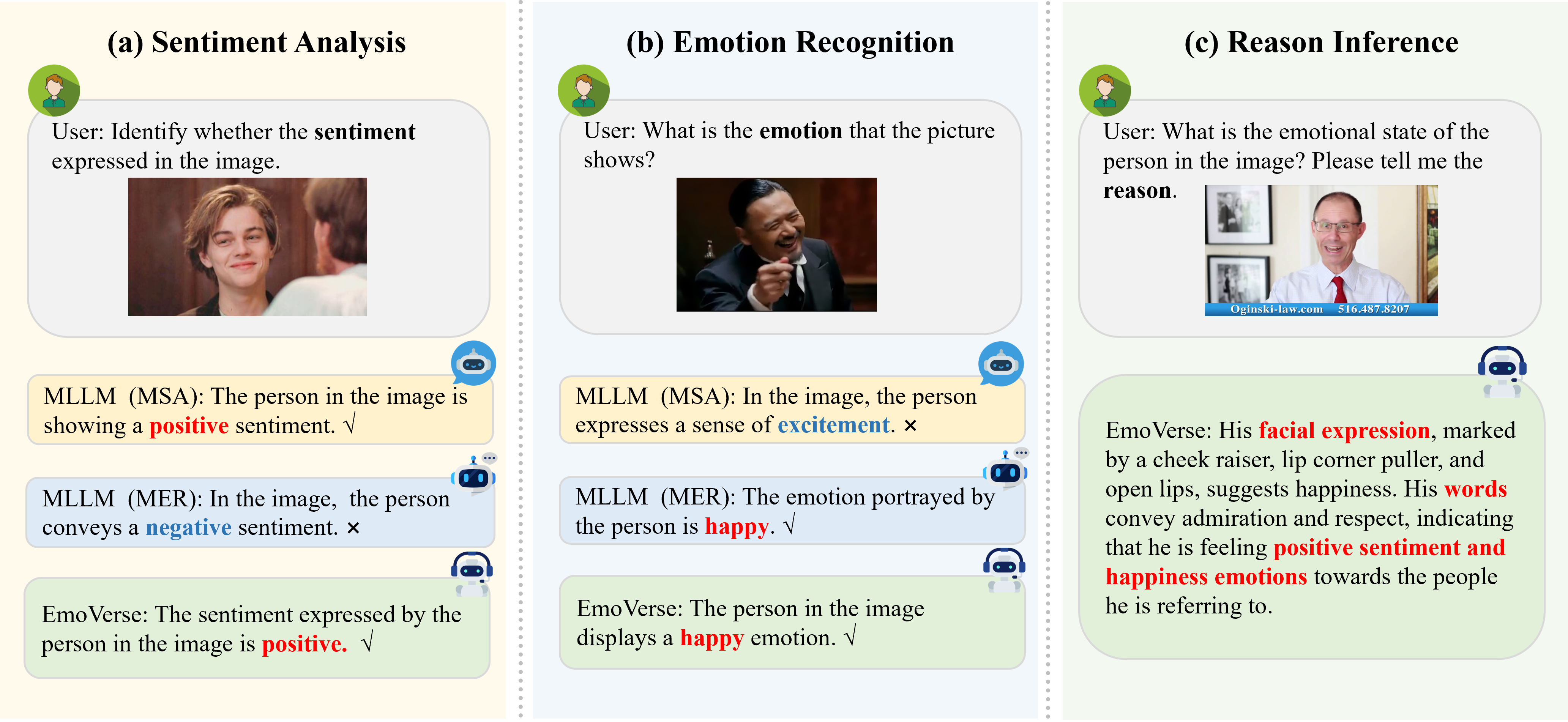}
  \caption{Comparison of MLLMs trained on the MSA task, MER task, and using the M2SE strategy. (a) The MSA-trained model performs well on MSA but poorly on MER. (b) The ER-trained model performs well on MER but poorly on MSA. (c) EmoVerse, trained with the M2SE strategy, excels on both MSA and MER tasks and successfully performs emotion reasoning inference.}
  \label{framework}
\end{figure*}

MLLMs can effectively perform multiple visual tasks in a zero-shot setting after visual instruction fine-tuning. 
However, in the field of affective computing, existing MLLMs continue to struggle with sentiment understanding after fine-tuning with emotion-related instruction data, as shown in Figure~\ref{framework} and exemplified by models such as Emotion-LLaMA~\cite{cheng2024emotion}. 
This discrepancy highlights a broader issue: \textbf{existing MLLMs fail to capture the intricate relationship between sentiment and emotion.} 
A promising solution to this challenge is multitask learning, where the model is trained simultaneously on multimodal sentiment analysis (MSA) and multimodal emotion recognition (MER) tasks. An example of this approach is UniMSE~\cite{hu2022unimse}, which demonstrates the effectiveness of multitask learning. However, a significant open challenge remains: how to optimally leverage multiple tasks to enhance MLLM training?

In response, we propose to train a general-purpose MLLM for affective computing using multitask learning. In addition to MSA and MER, we incorporate facial expression recognition (FER), emotion reasoning inference (ERI), and emotion cause-pair extraction (ECPE) tasks, aiming to improve the model's capabilities in facial expression analysis, causal reasoning, and long-context comprehension. The ECPE task requires the model to identify the emotion in a specific sentence within a dialogue and determine the cause of that emotion, which is linked to a previous sentence with multimodal setting. This task demands a high level of reasoning and context understanding from the model. By examining the relationships among these tasks, we propose an optimal training strategy called \textbf{M}ultistage \textbf{M}ultitask \textbf{S}entiment and \textbf{E}motion Instruction Tuning Strategy (M2SE), which is designed to enable MLLMs to achieve strong performance across a variety of affective tasks. This training strategy is adaptable to different MLLMs. Based on the M2SE strategy, we develop a MLLM named \textbf{Emotion Universe (EmoVerse)}.

Additionally, existing datasets are often insufficient for training MLLMs in multitask settings due to their lack of diversity and task coverage. To address this issue, we construct the \textbf{Affective Multitask (AMT) dataset}, which includes the five tasks mentioned above. By incorporating these tasks, the AMT dataset facilitates the development of models capable of recognizing, reasoning, and inferring the causes of sentiment and emotion.

In summary, our main contributions are as follows:
\begin{itemize}
    \item We construct the AMT dataset. Each piece of data in this dataset contains queries for five tasks and the corresponding labels. The AMT dataset encourages the model to capture the relationship between sentiment and emotion from the perspective of different tasks, enabling the model to extract richer contextual information.
    \item We explore the relationships between different sentiment and emotion-related tasks within a multitask learning framework. Through this exploration, we identify the optimal strategy, called M2SE. This exploration is the first multitask investigation in the field of affective computing.
    \item We develop the EmoVerse model. EmoVerse unifies tasks in the sentiment and emotion domains by leveraging the M2SE strategy. 
    \item EmoVerse demonstrates outstanding performance across a wide range of tasks. On the CMU-MOSEI dataset for MSA\cite{zadeh2018multimodal}, it achieves an Acc2 score of 88.51\%. Similarly, on the MELD dataset for MER\cite{poria2018meld}, it attains a weighted F1 score of 66.74.
\end{itemize}

\begin{figure*}[h]
  \centering
  \includegraphics[width=1\linewidth]{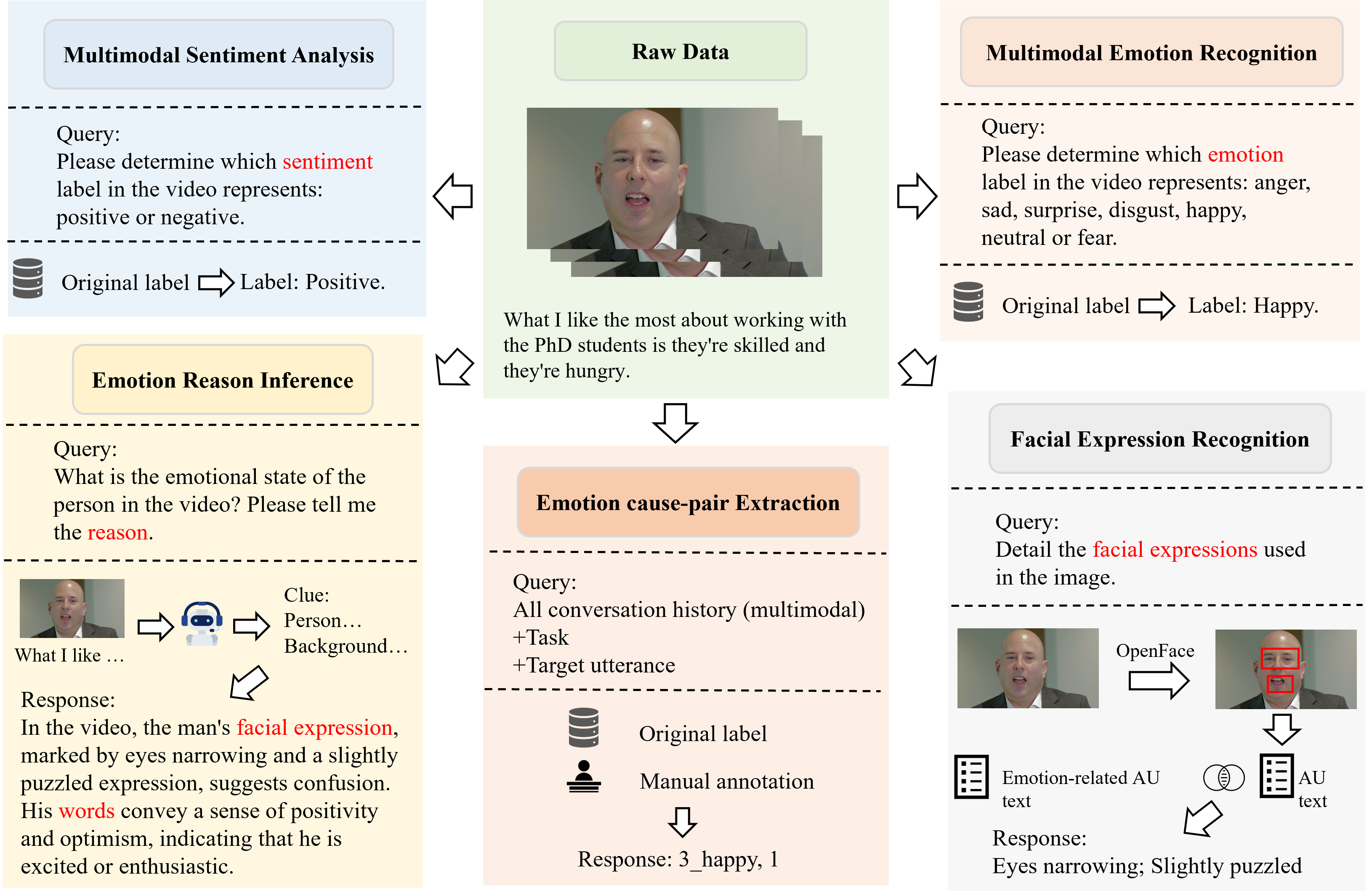}
  \caption{The construction process of the AMT dataset. The AMT dataset includes five tasks: multimodal sentiment analysis, multimodal emotion recognition, emotion reason inference, facial expression recognition, and emotion cause-pair extraction.}
  \label{amt}
\end{figure*}

\section{Related Work}

\subsection{Multimodal Large Language Models}
Multimodal Large Language Models (MLLMs) have gained significant attention due to their powerful reasoning capabilities. These models are typically trained for general-purpose tasks using a standard framework that includes modal encoders, a large language model (LLM), and a mapping layer (e.g., Linear \cite{chen2023minigpt}, MLP \cite{liu2024visual,chen2024far}, or Q-Former \cite{li2023blip}). While they excel in general tasks, MLLMs struggle with MSA and MER due to insufficient training on sentiment-specific datasets and emotion-related knowledge. Recent efforts have focused on improving MLLMs by incorporating multimodal emotion datasets and emotion reasoning tasks \cite{cheng2024emotion,lian2024explainablemultimodalemotionrecognition}. However, these models still face challenges in leveraging contextual information effectively, particularly in sentiment analysis, where understanding the  nuances is crucial.

\subsection{Multitask Learning}
Multitask Learning (MTL) is a machine learning paradigm that enables models to simultaneously learn multiple related tasks, thereby allowing knowledge from one task to be transferred to others. In affective computing, where multimodal inputs (e.g., text, audio, and visual frames) carry diverse and distinct information, MTL frameworks aim to leverage the synergies between tasks for improved performance. For instance, UniMSE \cite{hu2022unimse} unifies MSA and MER within a single framework. However, it relies solely on labeled data to establish the relationship between sentiment and emotion, without fully utilizing multimodal connections. M2Seq2Seq \cite{10.1016/j.inffus.2023.01.005} proposes a multimodal multitask learning model based on the encoder-decoder architecture. Recently, Emotion-LLaMA \cite{cheng2024emotion} has shown promising results in MER by incorporating multitask learning, including emotion cause inference. However, despite its success in MER, Emotion-LLaMA still struggles with MSA tasks and fails to effectively synthesize contextual information for emotion inference.

\begin{figure*}[h]
  \centering
  \includegraphics[width=1\linewidth]{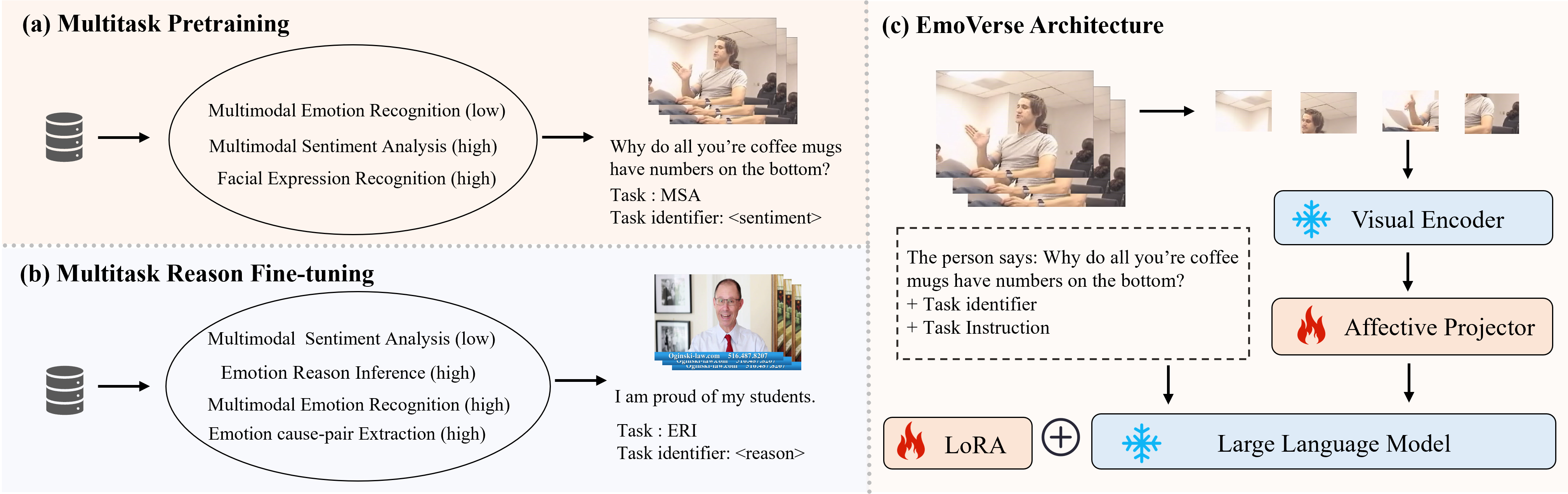}
  \caption{Overview of the M2SE training strategy and EmoVerse architecture. For different training stages, we assign tasks to the data based on their respective sampling rates. The EmoVerse model is then trained using the assigned data. The model architecture include visual encoder, linear projector, and LLM.}
  \label{m2se}
\end{figure*}

\section{Affective Multitask Dataset}
The AMT dataset is constructed by integrating data from the CMU-MOSEI, MELD, and ECF2.0~\cite{wang-etal-2024-semeval} datasets to support five distinct tasks: MSA, MER, FER, ERI, and ECPC. Figure~\ref{amt} illustrates a sample from the AMT dataset.

For the MSA and MER tasks, we use the original sentiment labels and emotion annotations from the source datasets as task labels. Due to the complexity of annotating the ECPE task, we refer to the annotation method of the ECF2.0 dataset and manually annotate a portion of the data. This is then combined with the existing ECF2.0 data to form the ECPE portion of the AMT dataset for further emotional causality analysis.

In the FER task, we adapt methods from the MERR dataset~\cite{cheng2024emotion} by extracting peak frames based on the Action Units (AUs) calculated for each frame using the OpenFace tool. The composite score for each frame is determined by summing the AU values, as shown in the following equation:

\begin{equation}
S_f = \sum_{i=1}^{n} \text{AU}_i(f)
\end{equation}

where \(S_f\) is the composite score of frame \(f\), and \(\text{AU}_i(f)\) represents the Action Unit value for the \(i\)-th facial action unit at frame \(f\). The peak frame with the highest score is selected to represent the most expressive moment. For videos featuring multiple characters, such as MELD dataset, we improve the method by extracting peak frames separately for each character and then selecting the final peak frame by comparing the highest composite scores from all characters:

\begin{equation}
S_{f_{\text{final}}} = \max \left( \{S_f^{(1)}, S_f^{(2)}, \dots, S_f^{(k)}\} \right)
\end{equation}

where \(S_f^{(i)}\) is the composite score for character \(i\) and \(k\) represents the total number of characters in the frame. The final peak frame is the one with the highest composite score among all characters.

After selecting the peak frames, we extract the facial expression captions by combining the AUs from the peak frame with the predefined AU list corresponding to the emotion. Let \( \text{AU}_\text{frame}(f) \) represent the set of AUs from the peak frame for a given emotion, and \( \text{AU}_\text{emotion}(e) \) represent the predefined AU list for a specific emotion \(e\). The common AUs are obtained by taking the intersection of the two sets:

\begin{equation}
    \text{AU}_\text{common}(f,e) = \text{AU}_\text{frame}(f) \cap \text{AU}_\text{emotion}(e)
\end{equation}

where \( \text{AU}_\text{common}(f,e) \) is the set of AUs common to both the peak frame and the emotion-specific AU list. These common AUs are then mapped to textual information to generate the FER annotations for the task. This approach enables the model to link the facial expressions captured in the peak frames with the corresponding emotional state.

In the ERI task, we input the video into the Internvl2-8B \footnote{\href{https://huggingface.co/OpenGVLab/InternVL2-8B}{https://huggingface.co/OpenGVLab/InternVL2-8B}} model to generate detailed descriptions of both the characters and the scene. These descriptions allow the model to analyze emotional causes by considering facial expressions, speech, and environmental context. Finally, we use the LLaMA 3.1-8B \footnote{\href{https://github.com/meta-llama/llama3}{https://github.com/meta-llama/llama3}} model to infer the underlying causes of the characters’ emotions by synthesizing the character and scene descriptions.

We conduct a comprehensive preprocessing and quality control procedure on the raw data from the three datasets, systematically removing problematic or noisy samples. In parallel, we manually curate high-quality samples according to the methodology described above. The final refined dataset, detailed in Table \ref{tab:amt_dataset}, includes the number of tasks per category. It is important to note that each data sample may be associated with a varying number of tasks. Table \ref{tab:amt_dataset} provides a detailed breakdown of the tasks available for each category.

Additionally, we transform the dataset into a structured format, which includes three key elements: query, response, and images/videos. This format facilitates the instruction tuning of MLLMs.

\begin{table}[H]
    \centering 
    \footnotesize
    \begin{tabularx}{\columnwidth}{
    >{\raggedright\arraybackslash}l
    >{\raggedleft\arraybackslash}X}
        \toprule
        \textbf{Tasks} & \textbf{Number of Entries} \\
        \midrule
        Multimodal Sentiment Analysis & 25859 \\
        Multimodal Emotion Recognition & 25859 \\
        Facial Expression Recognition & 15870 \\
        Emotion Reason Inference & 4839 \\
        Emotion Cause-Pair Extraction & 7081 \\
        \midrule
        Total & 32940 \\
        \bottomrule
    \end{tabularx}

    \caption{The Number of Entrie in the AMT Dataset}
    \label{tab:amt_dataset}
\end{table}

\section{Exploring the MLLM design space: How do different tasks influence the performance}
In this section, we explore how different tasks and varying sampling ratios for each task impact the final performance of the MLLM. After identifying the optimal training strategy, we introduce EmoVerse. 

Generally, the training process for MLLMs is divided into two stages: pretraining and instruction tuning. Inspired by the human learning process, which progresses from simple to complex, we also divide the affective domain fine-tuning process into these two stages, adopting a training approach that goes from simple to difficult. We conduct experiments using EmoVerse-4B as well as the AMT dataset (refer to Section 5.2).

In the AMT dataset, the simplest task is MSA, which requires the model to perform binary classification. The more challenging tasks are MER and FER, while ERI and ECPE, which involve reasoning and inference, are the most difficult. Based on this, our approach is to first train the model using MSA, MER, and FER in a multitask learning setup, with MSA receiving the highest sampling ratio due to its simplicity. In the second stage, we focus on training the model with MER, ERI, and ECPE in a multitask learning setup, building on the model's performance from the previous stage.

\subsection{Multitask Pretraining}

In this phase, we randomly select 15,000 samples from the AMT dataset. The primary objective is to enable the model to recognize facial expressions and, through these expressions, capture the relationship between sentiment and emotion. As discussed earlier, MSA is the simplest task, and therefore, it is assigned the highest sampling ratio, while MER, being a more complex task, receives a lower sampling ratio. In this stage, we also investigate the impact of varying the sampling ratio for the FER task on the model's performance. To assess the model's performance during this phase, we use the CMU-MOSEI test set.

\begin{table}[H]
    \centering
    \begin{adjustbox}{max width=\columnwidth}
    \begin{tabular}{>{\centering\arraybackslash}m{3cm}>{\centering\arraybackslash}m{3cm}}
        \toprule
        \textbf{Model} & \textbf{Acc2} \\
        \midrule
        EmoVerse (6:1:3) & 77.09 / 87.47  \\
        EmoVerse (6:3:1) & \textbf{82.61} / \textbf{87.50} \\
        \bottomrule
    \end{tabular}
    \end{adjustbox}
    \caption{Effect of the FER Task. MSA:FER:MER}
    \label{tab:fer}
\end{table}

As shown in Table \ref{tab:fer}, increasing the sampling rate for the FER task leads to improved performance on the MSA task. Consequently, for the first stage, we implement a multitask training setup with a sampling ratio of MSA:FER:MER = 6:3:1. This configuration enables the model to not only understand the relationship between sentiment and emotion but also accurately recognize facial action units (AUs), thereby enhancing MSA performance.

\subsection{Multitask Reason Fine-Tuning}

For this phase, we randomly select the remaining samples from the AMT dataset and distribute them across the tasks. As previously mentioned, in the second phase, we introduce more challenging tasks, including MER, ERI, and ECPE. The final performance of the MSA task is evaluated on the CMU-MOSEI dataset, the MER task performance on the MELD dataset, and the ECPE task performance on the ECF2.0 dataset (refer to Section 5.1). We begin with the MER task and gradually introduce additional tasks to examine the impact of training with different tasks. The results are summarized in Tabel~\ref{tab:dtasks}.

\begin{table}[ht]
    \centering
    \begin{adjustbox}{max width=\columnwidth}
    \begin{tabular}{lccccc}
        \toprule
        \multirow{2}{*}{\textbf{Model}} & \textbf{CMU-MOSEI} & \multicolumn{2}{c}{\textbf{MELD}} & \multicolumn{2}{c}{\textbf{ECF2.0}} \\
        \cmidrule(lr){2-6}
         & \textbf{Acc2} & \textbf{Acc} & \textbf{W.F1} & \textbf{F1} & \textbf{W.F1} \\
        \midrule
        EmoVerse (First stage) & 82.61 / 87.50 & - & - & - & - \\
        \midrule
        EmoVerse (MER) & 81.68 / 87.20 & 66.74 & 66.24 & 25.40 & 26.69 \\
        EmoVerse (MER,ERI,EPCE) & 79.00 / 86.78 & 66.97 & 65.67 & 69.85 & 69.95 \\
        EmoVerse (MER,ERI,EPCE,MSA) & \textbf{85.51} / \textbf{87.80}  & \textbf{67.78} & \textbf{66.74} & \textbf{72.30} & \textbf{72.21} \\
        \bottomrule
    \end{tabular}
    \end{adjustbox}
    \caption{Effect of the Distribution of Tasks on EmoVerse.}
    \label{tab:dtasks}
\end{table}

When we train the model in the second phase using only these three tasks, the final model performs poorly on the MSA task. This indicates the presence of catastrophic forgetting in MLLM multitask training. However, when we allocate a small portion of the MSA task in the second phase, overall performance improves. This suggests that revisiting MSA-related knowledge helps the model deepen its understanding of the relationship between sentiment and emotion. In the second phase, the final sampling ratio we chose is MER:ERI:ECPE:MSA = 3:3:3:1.

Therefore, the optimal training strategy we have identified involves using the MSA, FER, and MER tasks in the first phase, and the MSA, MER, ERI, and ECPE tasks in the second phase for multi-stage multitask training. We refer to this strategy as M2SE.

\subsection{EmoVerse}

To validate the effectiveness of the M2SE strategy, we select the most widely used MLLM framework without introducing any multitask modifications to its architecture. This ensures that any observed performance improvements can be directly attributed to the proposed training strategy, rather than architectural changes. The proposed MLLM architecture, EmoVerse, as shown in Figure \ref{m2se}, consists of a visual encoder, a linear projector, and the LLM.

We train EmoVerse following M2SE strategy using the AMT datasets. For a data sample \( D \), the input consists of the pair \( \{v, t\} \), where \( v \) represents the video and \( t \) represents the text. First, the video \( v \) is processed to extract visual tokens using a Vision Transformer (ViT)\cite{dosovitskiy2021imageworth16x16words}, producing a set of visual tokens \( T_v \). These visual tokens \( T_v \) are then passed through a linear transformation \( \mathbf{W}_v \) to align the visual features with the feature space of the LLM. Mathematically, this can be expressed as:
\begin{equation}
    T'_v = \mathbf{W}_v T_v
\end{equation}

where \( T'_v \) represents the aligned visual features.

Next, the text \( t \), task identifiers, and instructions are tokenized and passed through the embedding layer of the LLM, producing the tokenized representation \( T_t \). The tokenized visual features \( T'_v \) and the tokenized text \( T_t \) are then concatenated to form the final input to the model:
\begin{equation}
    \mathbf{X} = [T'_v; T_t]\
\end{equation}

where \( \mathbf{X} \) is the combined input. The model processes this concatenated input and outputs the response \( \hat{y} \), which corresponds to the task at hand:
\begin{equation}
    \hat{y} = \mathcal{M}(\mathbf{X})
\end{equation}

where \( \mathcal{M} \) is the model and \( \hat{y} \) is the model’s prediction or answer.

\begin{table*}[ht]
    \centering
    \footnotesize
        \begin{tabularx}{\textwidth}{
        >{\raggedright\arraybackslash}X
        >{\centering\arraybackslash}X
        >{\centering\arraybackslash}X
        >{\centering\arraybackslash}X
        >{\centering\arraybackslash}X
        >{\centering\arraybackslash}X
        >{\centering\arraybackslash}X}
        \toprule
        \multirow{2}*{Method}& \multicolumn{2}{c}{CMU-MOSEI} & \multicolumn{2}{c}{MELD} & \multicolumn{2}{c}{ECF2.0} \\
        \cmidrule(lr){2-7}
         & Acc2(N/N) & Acc2(N/P) & Acc & WF1 & F1 & WF1 \\
        \midrule
        MulT  & -  & 82.50 & - & - & - & -  \\
        Self-MM  & 82.81  &85.17 & - & - & - & -  \\
        MMIM  & 82.24  & 85.97 & - & - & - & -  \\
        BBFN  & -  & 86.20 & - & - & - & -  \\
        FDMER  & -  & 86.10 & - & - & - & -  \\
        MISA  & 83.60 & 85.50 & - & - & - & -  \\
        ConKI & 82.73  &86.25 & - & - & - & -  \\
        ConFEDE & 81.65 & 85.82 & - & - & - & -  \\
        PMR & - & 83.30 & - & - & - & -  \\
        MRC-D3AE & 83.10 & 85.50 & - & - & - & -  \\
        \midrule
        DialogueGCN & - & - & 59.46 & 58.10 & - & - \\
        DialogueCRN & - & - & 60.73 & 58.39 & - & - \\
        DAG-ERC     & - & - & -     & 63.65 & - & - \\
        MM-DFN       & - & - & -    & 59.46 & - & - \\
        EmoCaps      & -  & - & -   & 64.00 & - & - \\
        DisGCN  & -  & - & -   & 64.22 & - & - \\
        MMGCN       & - & - & 62.49 & 58.65 & - & - \\
        GA2MIF       & -  & - & -   & 58.94 & - & - \\
        UniMSE      & - & - & 65.09 & 65.51 & - & - \\
        FacialMMT    & -   & - & -  & 66.58 & - & - \\
        \midrule
        LLaVA-Video  & - & 82.82 & 23.87 & 22.65 & 22.59 & 22.78 \\
        InternVL2-4B & - & 82.09 & 39.46 & 40.59 & 33.24 & 33.35 \\
        InternVL2-8B & - & 85.30 & 42.18 & 44.13 & 40.57 & 40.51 \\
        \midrule
        EmoVerse-4B & 85.51 & 87.80 & 67.28 & 66.58 & 72.30 & 72.21 \\
        EmoVerse-8B & \textbf{85.93} & \textbf{88.51} & \textbf{67.78} & \textbf{66.74} & \textbf{73.54} & \textbf{73.62} \\
        \bottomrule
        \end{tabularx}
    \caption{Comparison of EmoVerse with other SOTA methods across different tasks.}
    \label{tab:result}
\end{table*}

\section{Experiments}
\subsection{Experiments Setup}
EmoVerse is trained solely on the AMT dataset without the addition of any external data, and it is tested on the following task-specific test sets.
\subsubsection{Sentiment Analysis} We use the CMU-MOSEI datasets for MSA. The CMU-MOSEI dataset is an extended version of the CMU-MOSI dataset. It encompasses 3228 videos from 1000 speakers, comprising a total of 23453 sentences. Similar to the CMU-MOSI dataset, it covers multiple emotional dimensions, employing the same annotation labeling approach. The results for Acc-2 calculated based on negative/non-negative (N/N) scheme, while the results on the right are calculated according to negative/positive (N/P) definition.

\subsubsection{Emotion Recognition} We use the MELD datasets for ER. The MELD dataset is an emotional dialogue dataset comprising 1433 dialogue segments and 13708 utterances from movies and TV shows. Each statement in the dialogue is labeled with one of seven emotions: anger, disgust, sadness, joy, neutral, surprise, and fear. Furthermore, the MELD data set provides annotations for emotional polarity (positive, negative, and neutral) for each utterance. We use Acc and weighted f1-sorce for evaluation.

\subsubsection{Emotion Cause-pair Extraction} 
We use the ECF2.0 datasets for ECPE. The ECF2.0 dataset is a comprehensive resource for emotion cause-pair extraction, comprising 1,715 conversations and 16,720 utterances. Each utterance in the dataset is annotated with emotion cause-pair information, which helps in understanding the emotional context and its causes. The dataset is divided into training and evaluation sets, with 1,374 conversations and 13,619 utterances in the training set, and 341 conversations and 3,101 utterances in the test set. We refer to previous evaluations and assess the ECF2.0 test set using the F1 Score and Weighted F1 Score. 

\begin{figure*}[h]
  \centering
  \includegraphics[width=1\linewidth]{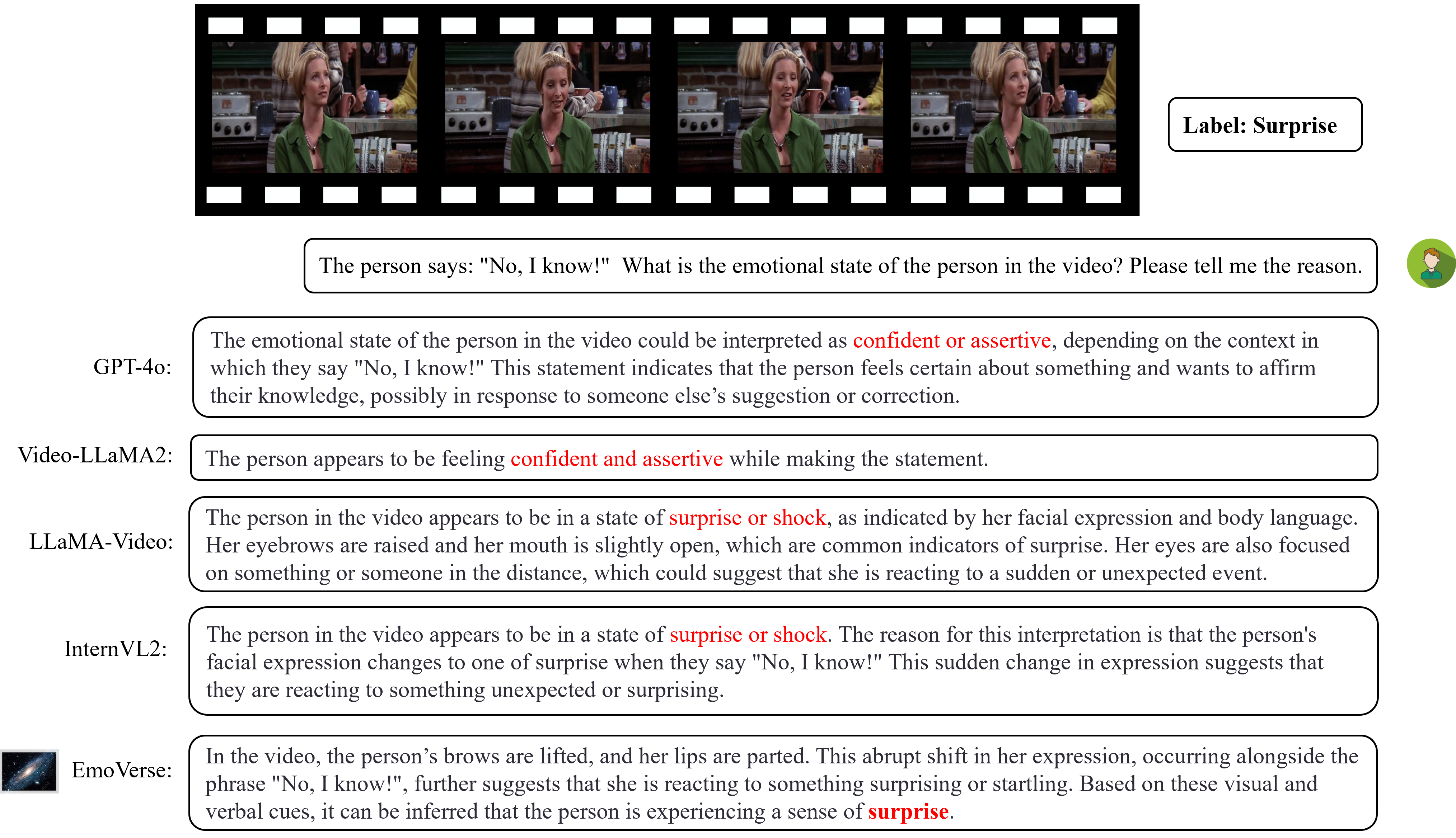}
  \caption{Comparison of different models on ERI. It can be seen that both GPT-4o and Video-LLaMA2 make errors in MER. While Llava-Video and InternVL2 correctly classify the emotions, their descriptions of the reasons behind the emotions are incomplete. In contrast, EmoVerse not only correctly classifies the emotions but also provides a comprehensive description of the reasons behind them.}
  \label{reason}
\end{figure*}

\subsection{Implementation Details}
For the visual encoder, we use a pre-trained ViT with input images resized to \( 448 \times 448 \) pixels. During fine-tuning, the visual backbone is frozen, and training focuses on the linear projection layer. For the LLM, EmoVerse-4B uses Phi-3-Mini \cite{abdin2024phi}, and EmoVerse-8B uses InternLM-2.5-7B-Chat\footnote{\url{https://github.com/InternLM/InternLM}}, both augmented with LoRA~\cite{hu2021lora} for parameter-efficient fine-tuning.

Training is done for 2 epochs with LoRA hyperparameters \( r = 8 \) and \( a = 32 \), a learning rate of \( 1 \times 10^{-5} \), and a warm-up cosine decay schedule. EmoVerse-4B training takes approximately 48 hours on 2 NVIDIA RTX 3090 GPUs, while EmoVerse-8B requires about 55 hours on a single NVIDIA A100 GPU.

\subsection{Results}
\subsubsection{Performance comparison} 

For MSA task, we compare our EmoVerse model with several SOTA baselines, including MulT \cite{tsai2019multimodal}, PMR \cite{lv2021progressive}, MISA \cite{hazarika2020misa}, MMIM \cite{han2021improving}, Self-MM \cite{yu2021learning}, BBFM \cite{han2021bi}, CONKI \cite{yu2023conki},  ConFEDE \cite{yang2023confede}, and MRC-D3AE \cite{zhang2024reconstructing}. In MER task, we evaluate our model against baseline models with DialogueGCN \cite{ghosal2019dialoguegcn}, DialogueCRN \cite{hu2021dialoguecrn}, DAG-ERC \cite{shen2021directed}, MM-DFN \cite{hu2022mm}, EmoCaps \cite{li2022emocaps}, DisGCN \cite{sun2021discourse}, MMGCN \cite{hu2021mmgcn}, GA2MIF \cite{li2023ga2mif}, UniMSE \cite{hu2022unimse}, and FacialMMT \cite{zheng2023facial}.

Additionally, for both MSA and MER,  we select models capable of processing both images and videos, including LLaVA-Video\cite{zhang2024video} and InternVL2\cite{chen2024internvlscalingvisionfoundation}. For the ECPE task, we conduct tests solely on MLLMs, as they are better equipped to handle the complexity and multimodal nature of the task. The performance comparison results, shown in Table \ref{tab:result} clearly demonstrate that EmoVerse achieves SOTA performance across both tasks, outperforming all the aforementioned baselines.

\subsubsection{Analysis of Emotion Reasoning} 
To demonstrate EmoVerse's qualitative performance, we compare emotion reasoning results across five models: GPT-4o, Video-LLaMA2\cite{damonlpsg2024videollama2}, LLaMA-Video, InternVL2, and EmoVerse, as shown in Figure \ref{reason}. The video depicts a woman displaying a strong surprise reaction, highlighting the models' ability to infer emotional states from multimodal inputs.

GPT-4o and Video-LLaMA2 misclassify the emotion as "confident" or "assertive," while LLaMA-Video and InternVL2 correctly identify it as "surprise." However, these models fail to fully integrate multimodal information for emotion reasoning. LLaMA-Video relies solely on facial expressions, while InternVL2 combines facial cues with textual analysis but lacks detailed explanation of the emotional cause. In contrast, EmoVerse effectively combines video and text to accurately infer the emotional cause, providing a comprehensive interpretation.

\begin{table}[H]
    \centering
    \begin{adjustbox}{max width=\columnwidth}
    \begin{tabular}{lccccc}
        \toprule
        \multirow{2}*{\textbf{Model}} & \textbf{CMU-MOSEI} & \multicolumn{2}{c}{\textbf{MELD}} & \multicolumn{2}{c}{\textbf{ECF2.0}} \\
        \cmidrule(lr){2-6}
         & \textbf{Acc2} & \textbf{Acc} & \textbf{W.F1} & \textbf{F1} & \textbf{W.F1} \\
        \midrule
        EmoVerse(Zero-shot) & 82.09 / 83.87 & 39.46 & 40.59 & 33.24 & 33.35 \\
        EmoVerse(MSA) & 83.96 / \textbf{88.16}  & 48.12 & 31.27 & 41.67 & 41.14 \\
        EmoVerse(MER) & 78.99 / 80.17 & 66.21 & 65.71 & 40.61 & 38.23 \\
        EmoVerse(EPCE) & 83.27 / 85.85  & 32.38 & 33.43 & \textbf{72.69} & \textbf{72.59} \\
        \midrule
        EmoVerse(M2SE) & \textbf{85.51} / 87.80  & \textbf{67.78} & \textbf{66.74} & 72.30 & 72.21 \\
        \bottomrule
    \end{tabular}
    \end{adjustbox}
    \caption{Comparison of the M2SE Strategy with Single-Task Fine-Tuning}
    \label{tab:m2se}
\end{table}

\subsection{Comparison of the M2SE Strategy with Single-Task Fine-Tuning}
To assess the impact of the M2SE strategy within the same framework and compare its performance across tasks, we conduct an ablation study. 
Table \ref{tab:m2se} lists the fine-tuning tasks (in parentheses) with their corresponding datasets. The M2SE strategy uses the AMT dataset for training. Results show that while some metrics may fall short of single-task learning, EmoVerse with M2SE  outperforms other models overall.

\section{Conclusion}
In this work, we explore the impact of different affective tasks on MLLM under a multistage and multitask training framework, identifying and validating the optimal strategy called M2SE. Using the M2SE strategy, we train EmoVerse, addressing the lack of a general-purpose MLLM in affective computing. Additionally, we construct the AMT dataset, designed to support sentiment and emotion perception training for MLLMs. Experimental results demonstrate that EmoVerse achieves state-of-the-art performance across multiple tasks. This study highlights the effectiveness of the M2SE strategy in unifying sentiment and emotion tasks, paving the way for more emotionally intelligent AI systems.

\section*{Acknowledgments}
This work was supported by the Key R\&D Program of Shandong Province, China (Major Scientific and Technological Innovation Project) (NO.2022CXGC010504) and National Natural Science Foundation of China under Grant 61301253.

\appendix

\bibliographystyle{named}
\bibliography{ijcai24.bib}

\end{document}